\documentclass[nopreprintline,12pt,authoryear]{elsarticle}

\usepackage{amssymb}

\usepackage{times}  
\usepackage{helvet}  
\usepackage{courier}  
\usepackage{url}  
\usepackage{graphicx}  

\usepackage{amssymb} 
\usepackage{pifont}
\usepackage{amsmath}
\usepackage{bm}
\usepackage{multirow}
\usepackage{subcaption}
\usepackage{color}

\newcommand{\cmark}{\ding{51}}%
\newcommand{\xmark}{\ding{55}}%
\newcommand{\vect}[1]{\bm{#1}}
\newcommand{\mat}[1]{\bf{#1}}
\newcommand{\RR}{\mathbb{R}}

\journal{Computer Speech \& Language}

\begin{document}

\begin{frontmatter}



\title{Sequential Neural Networks for Noetic End-to-End Response Selection}

\author{Qian Chen, Wen Wang \\
Speech Lab, DAMO Academy, Alibaba Group \\
\{tanqing.cq,~w.wang\}@alibaba-inc.com }

\author{}

\address{}

\begin{abstract}
The noetic end-to-end response selection challenge as one track in the 7th Dialog System Technology Challenges (DSTC7) aims to push the state of the art of utterance classification for real world goal-oriented dialog systems, for which participants need to select the correct next utterances from a set of candidates for the multi-turn context. This paper presents our systems that are ranked top 1 on both datasets under this challenge, one focused and small (Advising) and the other more diverse and large (Ubuntu). Previous state-of-the-art models use hierarchy-based (utterance-level and token-level) neural networks to explicitly model the interactions among different turns' utterances for context modeling. In this paper, we investigate a sequential matching model based only on chain sequence for multi-turn response selection. Our results demonstrate that the potentials of sequential matching approaches have not yet been fully exploited in the past for multi-turn response selection.  In addition to ranking top 1 in the challenge, the proposed model outperforms all previous models, including state-of-the-art hierarchy-based models, on two large-scale public multi-turn response selection benchmark datasets.

\end{abstract}



\end{frontmatter}



\section{Introduction}

Dialogue systems are gaining more and more attention due to their encouraging potentials and commercial values. With the recent success of deep learning models~\citep{DBLP:conf/aaai/SerbanSBCP16}, building an end-to-end dialogue system became feasible. However, building end-to-end multi-turn dialogue systems is still quite challenging, requiring the 
system to memorize and comprehend multi-turn
conversation context, rather than only considering the current utterance as in single-turn dialogue systems.

Multi-turn dialogue modeling can be divided into generation-based methods \citep{DBLP:conf/aaai/SerbanSBCP16,DBLP:conf/aaai/ZhouLCLCH17} and retrieval-based methods~\citep{DBLP:conf/sigdial/LowePSP15,DBLP:conf/acl/WuWXZL17}. The latter is the focus of the noetic end-to-end response selection challenge in the 7th Dialogue System Technology Challenges (DSTC7)\footnote{http://workshop.colips.org/dstc7/} \citep{DSTC7}. Retrieval-based methods select the best response from a candidate pool for the multi-turn context, which can be considered as performing a multi-turn response selection task.
The typical approaches for multi-turn response selection mainly consist of sequence-based methods~\citep{DBLP:conf/sigdial/LowePSP15,DBLP:conf/sigir/YanSW16} and hierarchy-based methods~\citep{DBLP:conf/emnlp/ZhouDWZYTLY16,DBLP:conf/acl/WuWXZL17,DBLP:conf/coling/ZhangLZZL18,DBLP:conf/acl/WuLCZDYZL18}. Sequence-based methods usually concatenate the context utterances into a long sequence. Hierarchy-based methods normally model each utterance individually and then explicitly model the interactions among the utterances. 

Recently, previous work \citep{DBLP:conf/acl/WuWXZL17,DBLP:conf/coling/ZhangLZZL18} claims that hierarchy-based methods with complicated networks can achieve significant gains over sequence-based methods. However, in this paper, we investigate the efficacy of a sequence-based method, i.e., Enhanced Sequential Inference Model (ESIM)~\citep{DBLP:conf/acl/ChenZLWJI17} originally developed for the natural language inference (NLI) task. Our systems are ranked top 1 on both datasets, i.e., Advising and Ubuntu datasets, under the DSTC7 response selection challenge. In addition, the proposed approach outperforms all previous models, including the previous state-of-the-art hierarchy-based methods, on two large-scale public benchmark datasets, the Lowe's Ubuntu~\citep{DBLP:conf/sigdial/LowePSP15} and
E-commerce datasets~\citep{DBLP:conf/coling/ZhangLZZL18}. 

Hierarchy-based methods mainly use extra neural
networks to explicitly model the multi-turn utterances' relationship. They also usually need to truncate the utterances in the multi-turn context to make them the same length and shorter than the maximum length. However, the lengths of different turns usually vary significantly in real tasks. When using a large maximum length, we need to add a lot of zero padding in hierarchy-based methods, which will increase computational complexity and memory cost drastically. When using a small maximum length, we may throw away some important information in the multi-turn context. We propose to use a sequence-based model, the ESIM model, in the multi-turn response selection task to effectively address the above-mentioned problem encountered by hierarchy-based methods. We concatenate the multi-turn context as a long sequence, and convert the multi-turn response selection task into a sentence pair binary classification task, i.e., whether the next sentence is the response for the current context. There are two major advantages of ESIM over hierarchy-based methods:
\begin{itemize}
\item First, since ESIM does not need to make each utterance the same length, it has less zero padding and hence could be more computationally efficient than hierarchy-based methods. 
\item Second, ESIM models the interactions between utterances in the context implicitly, yet in an effective way as described in the model description section, without 
using extra complicated networks.
\end{itemize}

This paper is an extended version of our paper~\citep{dstc19task1chen} presented at the DSTC7 workshop. It includes complete ablation analysis of various approaches we explored for our proposed system, and results and ablation analysis of exploring the Bidirectional Encoder Representations from Transformers (BERT) model~\citep{DBLP:journals/corr/abs-1810-04805} for the noetic end-to-end response selection task. The contribution of this paper can be summarized as follows.
\begin{enumerate}
    \item We develop an Enhanced Sequential Inference Model (ESIM) based system for the DSTC7 noetic end-to-end response selection track. On top of the ESIM model, we explore methods for exploiting multiple word embeddings, heuristic data augmentation, tuning the ratio between positive and negative samples, and emphasizing the importance of the most recent context utterances.
    \item We propose a two-step approach for selecting the next utterance from a large amount of candidates (i.e., for subtask 2 on the Ubuntu dataset, we need to select the next utterance from a candidate pool of 120,000 sentences), by first using a sentence-encoding based method to select the top N candidates from the large set of candidates and then reranking them using ESIM, achieving a high performance with an acceptable overall computational cost.
    \item We conduct systematic ablation analysis of the above-mentioned methods for enhancing the ESIM model performance. In particular, we develop effective and efficient model ensemble by averaging the output from models trained with different parameter initializations and different structures.
    Also particularly, we explore task-specific word embeddings for incorporating external domain knowledge in ESIM. Note that Subtask 5 of the DSTC7 noetic end-to-end response selection track was particularly designed to give access to external domain data, and to encourage the participants to explore how to effectively use the external information to boost the performance, as an effective use of external domain data for response selection remains a challenging task.
    \item Our final submitted system achieves the best performance overall on both Ubuntu and Advising datasets of the DSTC7 noetic end-to-end response selection track. We also demonstrate that the final system outperforms all previous models, including state-of-the-art hierarchy-based models, and achieves new state-of-the-art performances on two large-scale public multi-turn response selection benchmark datasets. Our source code is available at https://github.com/alibaba/esim-response-selection.
    \item We implement a BERT-based model for DSTC7 response selection track, observe significant improvement over our submitted system, conduct systematic ablation analysis for the BERT model, and compare the computational cost between the ESIM model and BERT model.
\end{enumerate}

\section{Task Description}

DSTC7 is divided into three different tracks, and the proposed approach is developed for the noetic end-to-end response selection track. This track focuses on goal-oriented multi-turn dialogues and the objective is to select the correct response from a set of candidates. Participating systems should not be based on hand-crafted features or rule-based systems. Two datasets are provided, i.e., Ubuntu and Advising, which will be introduced in detail in the experiment section.

\begin{figure}[!t]
\centering
\includegraphics[width=\textwidth]{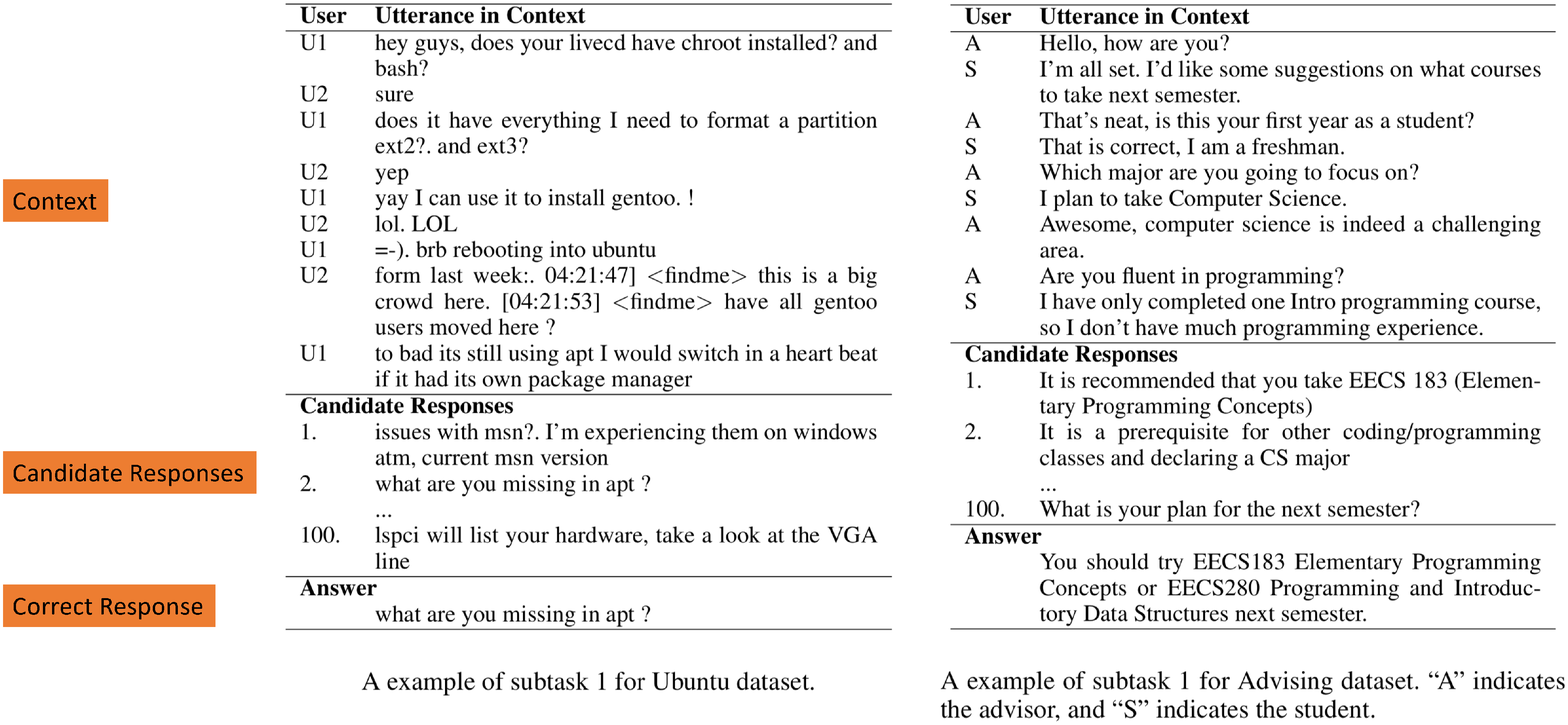}
\caption{Examples of the context, the candidate responses, and the correct response for the subtask 1 for the Ubuntu and Advising datasets, respectively.}
\label{fig:dstc7:example}
\end{figure}

The response selection track provided series of subtasks that have similar structures, but vary in the output space and available context. In Table~\ref{tab:stat:task}, \cmark~indicates that the task is evaluated on the marked dataset, and \xmark~indicates not applicable. Figure~\ref{fig:dstc7:example} shows examples of the context, the candidate responses, and the correct response for the subtask 1 for the Ubuntu and Advising datasets, respectively.

\begin{table*}[ht]
\begin{center}
\scalebox{0.9}{
\begin{tabular}{c p{9cm} c c}
\hline
\multicolumn{1}{l}{\textbf{Subtask}} & \multicolumn{1}{l}{\textbf{Description}} & \multicolumn{1}{l}{\textbf{Ubuntu}}  & \multicolumn{1}{l}{\textbf{Advising}}  \\
\hline
1 &Select the next utterance from a candidate pool of 100 sentences	  &\cmark & \cmark\\
2 &Select the next utterance from a candidate pool of 120000 sentences &\cmark & \xmark\\
3 &Select the next utterance and its paraphrases from a candidate pool of 100 sentences &\xmark & \cmark \\
4 &Select the next utterance from a candidate pool of 100 which might not contain the correct next utterance & \cmark & \cmark\\
5 & Select the next utterance from a candidate pool of 100 incorporating the external knowledge & \cmark & \cmark\\
\hline
\end{tabular}
}
\end{center}
\caption{The subtask descriptions for the Ubuntu and Advising datasets of the DSTC7 noetic end-to-end response selection track.}
\label{tab:stat:task}
\end{table*}

\section{Model Description}

\begin{figure*}[!t]
\centering
\begin{subfigure}[b]{0.7\textwidth}
\centering
\includegraphics[width=\textwidth]{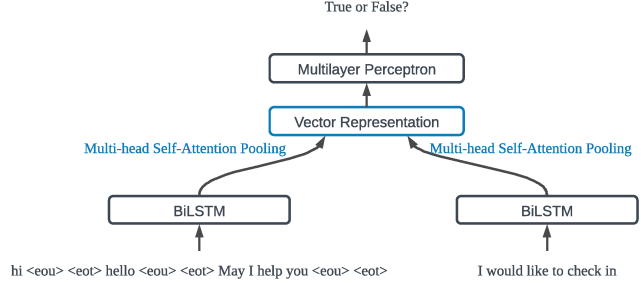}
\caption{Sentence-encoding based method.}
\label{fig:sent_enc}
\end{subfigure}
~ 
\begin{subfigure}[b]{0.7\textwidth}
\centering
\includegraphics[width=\textwidth]{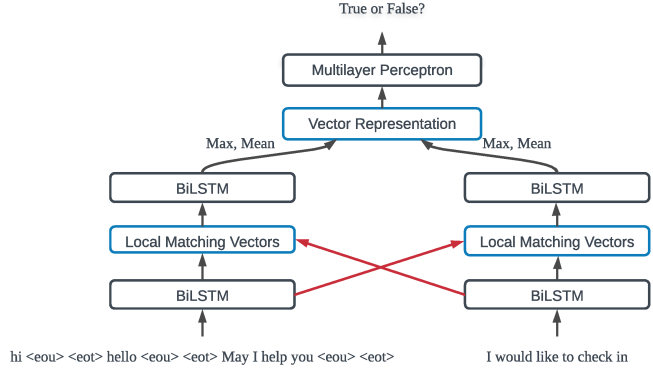}
\caption{Cross-attention based method.}
\label{fig:cross_att}	
\end{subfigure}
\caption{Two kinds of neural network based methods for sentence pair classification.}
\label{fig:fig_model}
\end{figure*}

The multi-turn response selection task is to select the next utterance from a candidate pool, given a multi-turn context. We convert the problem into a binary classification task, similar to the previous work~\citep{DBLP:conf/sigdial/LowePSP15,DBLP:conf/acl/WuWXZL17}. Given a multi-turn context and a candidate response, our model needs to determine whether or not the candidate response is the correct next utterance. In this section, we will introduce our model, Enhanced Sequential Inference Model (ESIM) \citep{DBLP:conf/acl/ChenZLWJI17} originally developed for natural language inference. The model consists of three main components, i.e., input encoding, local matching, and matching composition, as shown in Figure~\ref{fig:fig_model}(b). Section~\ref{sec:input} to Section~\ref{sec:matching-composition} basically recapitulate the ESIM model \citep{DBLP:conf/acl/ChenZLWJI17} with the modification of exploring multiple word embeddings described in Section~\ref{sec:input}.

\subsection{Input Encoding}
\label{sec:input}
Input encoding encodes the context information and represents tokens in their contextual meanings. Instead of encoding the context information through complicated hierarchical structures as in hierarchy-based methods, ESIM encodes the context information simply as follows. The multi-turn context is concatenated as a long sequence, which is denoted as $\vect{c} = (c_1,\dots,c_m)$, where $m$ is the total number of tokens in the multi-turn context. The candidate response is denoted as $\vect{r} = (r_1,\dots,r_{n})$, where $n$ is the number of tokens in the candidate response. Pretrained word embedding ${\mat E} \in \RR^{d_e \times |V|}$ is then used to convert $\vect{c}$ and $\vect{r}$ to two vector sequences $[{\mat E}(c_1),\dots,{\mat E}(c_m)]$ and $[{\mat E}(r_1),\dots,{\mat E}(r_n)]$, where $|V|$ is the vocabulary size and $d_e$ is the dimension of the word embedding. 

There are many kinds of pretrained word embeddings available, such as GloVe \citep{DBLP:conf/emnlp/PenningtonSM14} and fastText \citep{DBLP:conf/lrec/MikolovGBPJ18}. Different from the original ESIM model \citep{DBLP:conf/acl/ChenZLWJI17}, we propose a method to exploit multiple embeddings. Given $k$ kinds of pretrained word embeddings ${\mat E}_1, \dots, {\mat E}_k$ with their corresponding dimensions as $d_{e_1}, \dots, d_{e_k}$, we concatenate all embeddings for the word $i$, i.e., 
\begin{equation}
\label{eq1}
{\mat E}(c_i) = [{\mat E}_1(c_i);\dots;{\mat E}_k(c_i)]
\end{equation}
\noindent Then we use a feed-forward layer with ReLU to reduce the dimension from $(d_{e_1}+\dots+d_{e_k})$ to $d_h$, where $d_h$ is the dimension of the $\mathrm{BiLSTM}_1$ in Equation~\ref{eq2} and Equation~\ref{eq3}.

To represent tokens in their contextual meanings, the context and the response are fed into BiLSTM encoders to obtain context-dependent hidden states ${\vect c}^s$ and ${\vect r}^s$:
\begin{equation}
\label{eq2}
{\vect c}^s_i =\mathrm{BiLSTM}_1({\mat E}(\vect{c}),i)
\end{equation}

\begin{equation}
\label{eq3}
{\vect r}^s_j =\mathrm{BiLSTM}_1({\mat E}(\vect{r}),j)
\end{equation}
\noindent where $i$ and $j$ indicate the $i$-th token in the context and the $j$-th token in the response, respectively.

\subsection{Local Matching}
Modeling the local semantic relation between a context and a response is the critical component for determining whether the response is the proper next utterance. For instance, a proper response usually relates to some keywords in the context, which can be captured by modeling the local semantic relation. 
Instead of directly encoding the context and the response as two dense vectors, we use the \textit{cross-attention mechanism} to align the tokens from the context and the response, and then calculate the semantic relation at the token level. The attention weight is calculated as:
\begin{equation}
e_{ij} = ({\vect c}^s_i)^\mathrm{T} {\vect r}^s_j
\label{eq:eij}
\end{equation}

Soft alignment is used to obtain the local relevance between the context and the response, which is calculated by the attention matrix ${\mat e} \in \RR^{m \times n}$  in Equation~(\ref{eq:eij}). Then for the hidden state of the $i$-th token in the context, i.e., ${\vect c}^s_i$, which already encodes the token itself and its contextual meaning, the relevant semantics in the candidate response is 
identified as a vector ${\vect c}^d_i$, called \textit{dual vector} here, which is a weighted combination of all the response's states, more specifically as shown in Equation~(\ref{eq:a_d}).

\begin{equation}
\label{eq:a_c}
\alpha_{ij}  = \frac{\exp(e_{ij})}{\sum_{k=1}^{n}\exp(e_{ik})}
\end{equation}

\begin{equation}
\label{eq:a_d}
~{\vect c}^d_i =\sum_{j=1}^{n}\alpha_{ij} {\vect r}^s_j 
\end{equation}

\begin{equation}
\label{eq:b_c}
\beta_{ij}  = \frac{\exp(e_{ij})}{\sum_{k=1}^{m}\exp(e_{kj})}
\end{equation}

\begin{equation}
\label{eq:b_d}
~{\vect r}^d_j =\sum_{i=1}^{m}\beta_{ij} {\vect c}^s_i
\end{equation}
\noindent where ${\vect \alpha} \in \RR^{m \times n}$ and ${\vect \beta} \in \RR^{m \times n}$ are the normalized attention weight matrices with respect to the $2$-axis and $1$-axis. The similar calculation is performed for the hidden state of each token in the response, i.e., ${\vect r}^s_j$, 
as in
Equation~(\ref{eq:b_d}) to obtain the dual vector ${\vect r}^d_j$. 

By comparing the vector pair $<{\vect c}_i^s, {\vect c}_i^d>$, we can model the token-level semantic relation between the aligned token pairs. The similar calculation is also applied for the vector pair $<{\vect r}_j^s,{\vect r}_j^d>$. 

Then we collect local matching information as follows:
\begin{equation}
\label{eq:infer1}
{\vect c}^l_i = F([{\vect c}^s_i;{\vect c}^d_i;{\vect c}^s_i - {\vect c}^d_i ;{\vect c}^s_i \odot {\vect c}^d_i])
\end{equation}

\begin{equation}
\label{eq:infer2}
{\vect r}^l_j = F([{\vect r}^s_j;{\vect r}^d_j;{\vect r}^s_j - {\vect r}^d_j; {\vect r}^s_j \odot {\vect r}^d_j])
\end{equation}
\noindent where the heuristic matching approach~\citep{DBLP:conf/acl/MouMLX0YJ16} with difference and element-wise product is used here to obtain the local matching vectors ${\vect c}^l_i$ and ${\vect r}^l_j$ for the context and the response, respectively.
$F$ is a one-layer feed-forward neural network with \textit{ReLU} activation to reduce the dimension. 

\subsection{Matching Composition}
\label{sec:matching-composition}
Matching composition is realized as follows. To determine whether the response is the next utterance for the current context, we explore a composition layer to compose the local matching vectors (${\vect c}^l$ and ${\vect r}^l$) collected above:
\begin{equation}
{\vect c}^v_i = \mathrm{BiLSTM}_2({\vect c}^l,i)
\end{equation}

\begin{equation}
{\vect r}^v_j = \mathrm{BiLSTM}_2({\vect r}^l,j)
\end{equation}
Again we use BiLSTMs as building blocks for the composition layer, but the role of BiLSTMs here is completely different from that in the input encoding layer. The BiLSTMs here read local matching vectors, ${\vect c}^l$ and ${\vect r}^l$, and learn to discriminate critical local matching vectors for the overall utterance-level relationship. 

The output hidden vectors of $\mathrm{BiLSTM}_2$ are converted to fixed-length vectors through pooling operations and fed to the final classifier to determine the overall relationship. Max and mean poolings are used and concatenated altogether to obtain a fixed-length vector. Then the final vector is fed to the multi-layer perceptron (MLP) classifier, with one hidden layer with \textit{tanh} activation, and \textit{softmax} output layer. The entire ESIM model is trained via minimizing the cross-entropy loss in an end-to-end manner.

\begin{equation}
y = \mathrm{MLP}([{\vect c}^v_{max};{\vect c}^v_{mean};{\vect r}^v_{max};{\vect r}^v_{mean}])
\end{equation}

\subsection{Sentence-encoding based Methods}
For subtask 2 on the Ubuntu dataset, we need to select the next utterance from a candidate pool of 120,000 sentences. If we use the cross-attention based ESIM model directly, the computational cost is unacceptable. Instead, we first use a sentence-encoding based method to select the top 100 candidates from 120,000 sentences and then rerank the top 100 candidates using ESIM. 

Sentence-encoding based models use the Siamese architecture~\citep{DBLP:conf/nips/BromleyGLSS93,DBLP:conf/repeval/ChenZLWJI17} shown in Figure~\ref{fig:fig_model} (a).  Parameter-tied neural networks are applied to encode both the context and the response. Then a neural network classifier is applied to decide the relationship between the two sentences. Here, we use BiLSTMs with multi-head self-attention pooling to encode sentences~\citep{DBLP:journals/corr/LinFSYXZB17,DBLP:conf/coling/ChenLZ18}, and an MLP to classify.

We use the same input encoding process as ESIM. To transform a variable length sentence into a fixed length vector representation, we use a weighted summation of all BiLSTM hidden vectors (${\mat H}$). The multi-head attention weight matrix ${\mat A}$ is computed as follows:
\begin{equation}
{\mat A} = \text{softmax}({\mat W}_2\text{ReLU}({\mat W}_1 {\mat H}^\mathrm{T} + {\vect b_1} ) + {\vect b_2})^\mathrm{T}
\end{equation} 
\noindent where 
\begin{itemize}
\item $d_a$ is the dimension of the attention network;
\item $d_h$ is the dimension of BiLSTMs;
\item $d_m$ is a hyperparameter of the head number that needs to be tuned using the development set;
\item ${\mat W}_1 \in \RR^{d_a \times 2d_h}$ and ${\mat W}_2 \in \RR^{d_m \times d_a}$ are weight matrices; 
\item ${\vect b_1} \in \RR^{d_a}$ and ${\vect b_2} \in \RR^{d_m}$ are bias; 
\item ${\mat H} \in \RR^{T \times 2d_h}$ are the hidden vectors of BiLSTMs, where $T$ denotes the length of the sequence;
\item ${\mat A} \in \RR^{T \times d_m}$ is the multi-head attention weight matrix. 
\end{itemize}

Instead of using max pooling or mean pooling, we sum up the BiLSTM hidden states ${\mat H}$ according to the weight matrix ${\mat A}$ to get a vector representation of the input sentence:
\begin{equation}
{\mat V} =  {\mat A}^\mathrm{T}{\mat H}
\end{equation} 
\noindent where the matrix ${\mat V} \in \RR^{d_m \times 2d_h}$ can be flattened into a vector representation ${\vect v} \in \RR^{2d_h d_m }$. 

To enhance the relationship between the sentence pairs, similarly to ESIM, we concatenate the embeddings of two sentences and their absolute difference and element-wise product~\citep{DBLP:conf/acl/MouMLX0YJ16} as the input to the MLP classifier:
\begin{equation}
y = \mathrm{MLP}([{\vect v}_c;{\vect v}_r;\lvert{\vect v}_c - {\vect v}_r\rvert; {\vect v}_c \odot {\vect v}_r])
\end{equation}
 
The MLP has two hidden layers with \textit{ReLU} activation, shortcut connections, and \textit{softmax} output layer. The entire model is trained end-to-end through minimizing the cross-entropy loss. 

\section{Experimental Results and Discussion}
In this section, we first introduce the datasets (Section~\ref{subsec:datasets}) and the model training details (Section~\ref{subsect:trainingdetails}). We then present the official DSTC7 evaluation metrics and the official results of our system and other systems (Section~\ref{subsect:officialresults}) , comparison of our proposed ESIM system with previous work on two large-scale public benchmarks (Section~\ref{subsect:comparison}), the ablation analysis for our proposed ESIM model (Section~\ref{subsect:ablationanalysis}),  and evaluations of our post-DSTC7 BERT models for response selection and their ablation analysis (Section~\ref{subsect:bert}).

\subsection{Datasets}
\label{subsec:datasets}
We evaluate our model on both datasets of the DSTC7 response selection track, i.e., the Ubuntu and Advising datasets. In addition, to compare with previous methods, we also evaluate our model on two large-scale public multi-turn response selection benchmarks, i.e., the Lowe's Ubuntu dataset~\citep{DBLP:conf/sigdial/LowePSP15} and the E-commerce dataset~\citep{DBLP:conf/coling/ZhangLZZL18}.
\subsubsection{Ubuntu Dataset}
The Ubuntu dataset includes two party conversations from Ubuntu Internet Relay Chat (IRC) channel~\citep{kummerfeld2018analyzing}. Under this challenge, the context of each dialogue contains more than 3 turns and the system is asked to select the next turn from the given set of candidate sentences. Linux manual pages are also provided as external domain knowledge. We use a similar heuristic data augmentation strategy as in \citep{DBLP:conf/sigdial/LowePSP15}, i.e., we consider each utterance (starting at the second utterance of a conversation) as a potential response, with the previous utterances as its context. Hence a dialogue of length 10 yields 9 training examples. To train a binary classifier, we need to sample negative responses from the candidate pool. Initially, we use a 1:1 ratio between positive and negative responses for balancing the samples. Later, we find using more negative responses improves the results, such as 1:4 or 1:9. Considering efficiency, we choose 1:4 in the final configuration for all subtasks except 1:1 for subtask 2.

\subsubsection{Advising Dataset}
The advising dataset includes two-party dialogues that simulate a discussion between a student and an academic advisor. Structured information is provided as a database including course information and personas. The data also includes paraphrases of the sentences and the target responses. We use a similar data augmentation strategy as for the Ubuntu dataset based on original dialogues and their paraphrases. The ratio between positive and negative responses is 1:4.33.

\subsubsection{Lowe's Ubuntu Dataset}
This dataset is similar to the DSTC7 Ubuntu data. The training set contains one million context-response pairs and the ratio between positive and negative responses is 1:1. On both development and test sets, each context is associated with one positive response and 9 negative responses. 

\subsubsection{E-commerce Dataset}
The E-commerce dataset~\citep{DBLP:conf/coling/ZhangLZZL18} is collected from real-word conversations between customers and customer service staff from Taobao\footnote{https://www.taobao.com}, the largest e-commerce platform in China. The ratio between positive and negative responses is 1:1 in both training and development sets, and 1:9 in the test set. 

\subsection{Training Details}
\label{subsect:trainingdetails}
We use spaCy\footnote{https://spacy.io/} to tokenize text for the two DSTC7 datasets, and use original tokenized text without any further pre-processing for the two public benchmark datasets. The multi-turn context is concatenated and two special tokens, \_\_eou\_\_ and \_\_eot\_\_, are inserted, where \_\_eou\_\_ denotes end-of-utterance and \_\_eot\_\_ denotes end-of-turn. 

The hyperparameters are tuned based on the development set.
We use GloVe \citep{DBLP:conf/emnlp/PenningtonSM14} and fastText~\citep{DBLP:conf/lrec/MikolovGBPJ18} as pretrained word embeddings. For subtask 5, we also use word2vec~\citep{DBLP:conf/nips/MikolovSCCD13} to train word embeddings from the provided Linux manual pages for the Ubuntu dataset, and to train word embeddings from the course information for the Advising dataset.
The statistics of the pretrained word embeddings are summarized in Table \ref{tab:stat:word}.

For Lowe's Ubuntu and E-commerce datasets, we use pretrained word embeddings on the training data by word2vec~\citep{DBLP:conf/nips/MikolovSCCD13}. 
The pretrained embeddings are fixed during the training procedure for the two DSTC7 datasets, but fine-tuned for Lowe's Ubuntu and E-commerce datasets. 

Adam~\citep{DBLP:journals/corr/KingmaB14} is used for optimization with an initial learning rate of $0.0002$ for Lowe's Ubuntu dataset, and $0.0004$ for the rest. The mini-batch size is set to $128$ for DSTC7 datasets, $16$ for the Lowe's Ubuntu dataset, and $32$ for the E-commerce dataset. The hidden size of BiLSTMs and MLP is set to 300.

To make the sequences shorter than the maximum length, we cut off last tokens for the response but do the cut-off in the reverse direction for the context, as we hypothesize that the last few utterances in the context is more important than the first few utterances.
For the Lowe's Ubuntu dataset, the maximum lengths of the context and the response are set to 400 and 150, respectively; for the E-commerce dataset, 300 and 50; for the DSTC7 datasets, 300 and 30. 

More specially, for subtask 2 of DSTC7 Ubuntu, we use BiLSTM hidden size 400 and 4 heads for sentence-encoding methods. For subtask 4, the candidate pool may not contain the correct next utterance, so we need to choose a threshold. When the probability of positive labels is smaller than the threshold, we predict that the candidate pool does not contain the correct next utterance. The threshold is selected from the range $[0.50,0.51,..,0.99]$ based on the development set. 

\begin{table}[ht]
\begin{center}
\scalebox{0.9}{
\begin{tabular}{l l r}
\hline
\multicolumn{1}{l}{\textbf{Embedding}} & \multicolumn{1}{l}{\textbf{Training corpus}} & \multicolumn{1}{l}{\textbf{\#Words}}   \\
\hline
glove.6B.300d & Wikipedia + Gigaword  & 0.4M \\
glove.840B.300d & Common Crawl   & 2.2M\\
glove.twitter.27B.200d & Twitter &  1.2M\\
wiki-news-300d-1M.vec & Wikipedia + UMBC  & 1.0M \\
crawl-300d-2M.vec & Common Crawl  &  2.0M \\
linux.word2vec.300d & Linux manual pages & 0.3M \\
course.word2vec.300d & Course information & 4K \\
\hline
\end{tabular}
}
\end{center}
\caption{Statistics of the pretrained word embeddings. Rows 1-3 are from GloVe; Rows 4-5 are from fastText; Rows 6-7 are from word2vec.}
\label{tab:stat:word}
\end{table}

\subsection{DSTC7 Results}
\label{subsect:officialresults}
Several different evaluation metrics are used for response selection. Recall@N is used by the challenge organizers following~\cite{DBLP:conf/sigdial/LowePSP15}, which counts how often the correct answer is within the top N specified by a system. For DSTC7 results in this paper, N is set to 1, 10, 50, due to the large candidate set (100 candidates)~\citep{dstc19task1Gunasekara}. Mean Reciprocal Rank (MRR) as a widely used metric from the ranking literature is also used by the challenge organizers~\citep{dstc19task1Gunasekara}.

Our official results on all DSTC7 response selection subtasks are summarized in Table~\ref{tab:result}. The challenge organizers consider the average of Recall@10 and MRR for ranking the teams. On the Advising dataset, the test case 2 (Advising2) results are considered for ranking, because test case 1 (Advising1) has some dependency on the training dataset. Subtask 3 may contain multiple correct responses, so Mean Average Precision (MAP) is considered as an extra metric.

To compare with other systems, Table~\ref{tab:result:dstc7} presents the official scores for each team which submitted results for all 8 subtasks of the DSTC7 response selection track. More details can be found in \citep{gunasekara2019dstc7}. Among 8 subtasks in total, our results (Team 3) rank top 1 on 7 subtasks, rank the second best on subtask 2 of Ubuntu, and overall rank top 1 on both datasets of the response selection challenge\footnote{The official evaluation allows up to 3 different settings, but we only submitted one setting.}. Team 10 did best on Subtask2 on Ubuntu, which used an extra TF-IDF based method to filter the 120,000 candidates into 100 options. 

\begin{table}[ht!]
\begin{center}
\scalebox{0.9}{
\begin{tabular}{l l r r r}
\hline
\multicolumn{1}{l}{\textbf{Subtask}} & \multicolumn{1}{l}{\textbf{Measure}} & \multicolumn{1}{l}{\textbf{Ubuntu}} & \multicolumn{1}{l}{\textbf{Advising1}} & \multicolumn{1}{l}{\textbf{Advising2}}  \\
\hline
\multirow{4}{*}{Subtask1} & Recall@1 & 0.645 & 0.398 & 0.214 \\
& Recall@10 & 0.902 & 0.844 & 0.630 \\
& Recall@50 & 0.994 & 0.986 & 0.948 \\
& MRR & 0.7350 & 0.5408 & 0.3390 \\
& \textbf{Metric} & 0.819 &  & 0.485 \\
\hline
\multirow{4}{*}{Subtask2} & Recall@1 & 0.067 & \multicolumn{2}{c}{\multirow{4}{*}{NA}} \\
& Recall@10 & 0.185 & & \\
& Recall@50 & 0.266 & & \\
& MRR & 0.1056 & & \\
& \textbf{Metric} & 0.145 & & \\
\hline
\multirow{5}{*}{Subtask3}& Recall@1 & \multirow{5}{*}{NA}  & 0.476 & 0.290 \\
& Recall@10 & & 0.906 & 0.750 \\
& Recall@50 & & 0.996 & 0.978 \\
& MRR & & 0.6238 & 0.4341 \\
& MAP & & 0.7794 & 0.5327 \\
& \textbf{Metric} &  & & 0.592  \\
\hline
\multirow{4}{*}{Subtask4} & Recall@1 & 0.624 & 0.372 & 0.232 \\
& Recall@10 & 0.941 & 0.886 & 0.692 \\
& Recall@50 & 0.997 & 0.990 & 0.938 \\
& MRR & 0.7420 & 0.5409 & 0.3826 \\
& \textbf{Metric} & 0.842 & & 0.537 \\
\hline
\multirow{4}{*}{Subtask5} & Recall@1 & 0.653 & 0.398 & 0.214 \\
& Recall@10 & 0.905 & 0.844 & 0.630 \\ 
& Recall@50 & 0.995 & 0.986 & 0.948 \\
& MRR & 0.7399 & 0.5408 & 0.3390 \\
& \textbf{Metric} & 0.822 & & 0.485 \\
\hline

\end{tabular}
}
\end{center}
\caption{The official submission results from our proposed ESIM system on the hidden test sets for the DSTC7 noetic end-to-end response selection challenge. NA - not applicable. The official metric used for ranking teams, denoted \textbf{Metric}, is the average of MRR and Recall@10, as presented in the table.}
\label{tab:result}
\end{table}

\begin{table}[ht]
\begin{center}
\scalebox{0.9}{
\begin{tabular}{l l r r r r r r r r}  
\hline
\multicolumn{1}{l}{\textbf{Team No.}} & \multicolumn{1}{l}{\textbf{Model}} & \multicolumn{4}{c}{\textbf{Ubuntu, Subtask}} & \multicolumn{4}{c}{\textbf{Advising2, Subtask}}  \\
& & 1 & 2 & 4 & 5 & 1 &3 &4 & 5 \\
\hline
2 & ESIM & 0.672 & 0.033 & 0.713 & 0.672 & 0.430 & 0.540 & 0.479 & 0.430 \\
10 & ESIM & 0.651 & \textbf{0.307} & 0.696 & 0.693 & 0.361 & 0.434 & 0.262 & 0.361 \\
18 & GRU+Att. & 0.690 & 0.000 & 0.721 & 0.710 & 0.287 & 0.380 & 0.398 & 0.326 \\
\hline
3 & ESIM & \textbf{0.819} & 0.145 & \textbf{0.842} &\textbf{ 0.822} & \textbf{0.485} &\textbf{0.592} & \textbf{0.537} & \textbf{0.485} \\
\hline
\end{tabular}
}
\end{center}
\caption{The official DSTC7 noetic end-to-end response selection track results cited from \citep{gunasekara2019dstc7}. Teams which submitted results for all subtasks are shown here. We are Team 3. The metric is the average of MRR and Recall@10.}
\label{tab:result:dstc7}
\end{table}

\begin{table*}[hbt!]
\begin{center}
\scalebox{0.85}{
\begin{tabular}{l  c c c  c c c}
\hline
\multicolumn{1}{l}{\textbf{Models}} & \multicolumn{3}{c}{\textbf{Ubuntu}} &  \multicolumn{3}{c}{\textbf{E-commerce}} \\
& \textbf{R@1} & \textbf{R@2}  & \textbf{R@5}  & \textbf{R@1}  & \textbf{R@2}  & \textbf{R@5}  \\
\hline
TF-IDF~\citep{DBLP:conf/sigdial/LowePSP15} & 0.410 & 0.545 & 0.708 & 0.159 & 0.256 & 0.477 \\
RNN~\citep{DBLP:conf/sigdial/LowePSP15} & 0.403 & 0.547 & 0.819 & 0.325 & 0.463 & 0.775 \\
CNN~\citep{DBLP:journals/corr/KadlecSK15} & 0.549 & 0.684 & 0.896 & 0.328 & 0.515 & 0.792 \\
LSTM~\citep{DBLP:journals/corr/KadlecSK15} & 0.638 & 0.784 & 0.949 & 0.365 & 0.536 & 0.828 \\
BiLSTM~\citep{DBLP:journals/corr/KadlecSK15} & 0.630 & 0.780 & 0.944 & 0.355 & 0.525 & 0.825 \\
\hline
MV-LSTM~\citep{DBLP:conf/ijcai/WanLXGPC16} & 0.653 & 0.804 & 0.946 & 0.412 & 0.591 & 0.857 \\
Match-LSTM~\citep{DBLP:conf/naacl/WangJ16} & 0.653 & 0.799 & 0.944 & 0.410 & 0.590 & 0.858 \\
Attentive-LSTM~\citep{DBLP:journals/corr/TanXZ15} & 0.633 & 0.789 & 0.943 &  0.401 & 0.581 & 0.849 \\
Multi-Channel~\citep{DBLP:conf/acl/WuWXZL17} & 0.656 & 0.809 & 0.942 & 0.422 & 0.609 & 0.871 \\
\hline
Multi-View~\citep{DBLP:conf/emnlp/ZhouDWZYTLY16} & 0.662 & 0.801 & 0.951 & 0.421 & 0.601 & 0.861 \\
DL2R~\citep{DBLP:conf/sigir/YanSW16} & 0.626 & 0.783 & 0.944 & 0.399 & 0.571 & 0.842 \\
SMN~\citep{DBLP:conf/acl/WuWXZL17} & 0.726 & 0.847 & 0.961 & 0.453 & 0.654 & 0.886 \\
DUA~\citep{DBLP:conf/coling/ZhangLZZL18} & 0.752 & 0.868 & 0.962 & 0.501 & 0.700 & 0.921 \\
DAM~\citep{DBLP:conf/acl/WuLCZDYZL18}  & 0.767 & 0.874 & 0.969  & - & - & - \\
\hline
\textbf{Our ESIM} & \textbf{0.796}& \textbf{0.894} & \textbf{0.975}   &  \textbf{0.570} & \textbf{0.767} & \textbf{0.948} \\
\hline
\end{tabular}
}
\end{center}
\caption{Comparisons of different models on two large-scale public benchmark datasets. All the results except ours are cited from the previous works~\citep{DBLP:conf/coling/ZhangLZZL18,DBLP:conf/acl/WuLCZDYZL18}.} 
\label{tab:result:public}
\end{table*}

\subsection{Comparison with Previous Work}
\label{subsect:comparison}
The results on the two large-scale public benchmarks are summarized in Table~\ref{tab:result:public}. The first group of models includes sentence-encoding based methods. They use hand-craft features or neural network features to encode both context and response, then a cosine classifier or MLP classifier is applied to decide the relationship between the two sequences. Previous work uses TF-IDF, RNN \citep{DBLP:conf/sigdial/LowePSP15}, CNN, LSTM, and BiLSTM \citep{DBLP:journals/corr/KadlecSK15} to encode the context and the response. 

The second group of models consists of sequence-based matching models, which usually use the attention mechanism, including MV-LSTM~\citep{DBLP:conf/ijcai/WanLXGPC16}, Matching-LSTM~\citep{DBLP:conf/naacl/WangJ16}, Attentive-LSTM~\citep{DBLP:journals/corr/TanXZ15}, and Multi-Channels~\citep{DBLP:conf/acl/WuWXZL17}. These models compare the token-level relationship between the context and the response, rather than comparing the two dense vectors directly as in sentence-encoding based methods. These kinds of models achieve significantly better performance than the first group of models. 

The third group of models includes more complicated hierarchy-based models, which usually model the token-level and utterance-level information explicitly. The Multi-View model~\citep{DBLP:conf/emnlp/ZhouDWZYTLY16} utilizes utterance relationships from the word sequence view and the utterance sequence view. The DL2R model~\citep{DBLP:conf/sigir/YanSW16} employs neural networks to reformulate the last utterance with other utterances in the context. The SMN model~\citep{DBLP:conf/acl/WuWXZL17} uses CNN and attention to match a response with each utterance in the context. The DUA model~\citep{DBLP:conf/coling/ZhangLZZL18} and the DAM model~\citep{DBLP:conf/acl/WuLCZDYZL18} apply a similar framework as SMN~\citep{DBLP:conf/acl/WuWXZL17}, where one improves with gated self attention and the other improves with the Transformer structure~\citep{DBLP:conf/nips/VaswaniSPUJGKP17}. 

Although the previous hierarchy-based work claimed that they achieved the state-of-the-art performance by using the hierarchical structure of the multi-turn context, our ESIM sequential matching model outperforms all previous models, including hierarchy-based models. On the Lowe's Ubuntu dataset, the ESIM model brings significant gains on performance over the previous best results from the DAM model, up to 79.6\% (from 76.7\%) R@1, 89.4\% (from 87.4\%) R@2 and 97.5\% (from 96.9\%) R@5. For the E-commerce dataset, the ESIM model also accomplishes substantial improvement over the previous state of the art by the DUA model, up to 57.0\% (from 50.1\%) R@1, 76.7\% (from 70.0\%) R@2 and 94.8\% (from 92.1\%) R@5. These results demonstrate the effectiveness of the ESIM model, a sequential matching method, for multi-turn response selection.

\subsection{Ablation Analysis}
\label{subsect:ablationanalysis}
\subsubsection{Effect of Context Composition, External Domain Knowledge, and Model Ensemble}

Table~\ref{tab:result:ubuntu} and~\ref{tab:result:advising} show the ablation analysis of context composition, emphasizing most recent context utterances, incorporating external domain knowledge, and model ensemble, for the DSTC7 Ubuntu and Advising datasets, respectively. We choose the highest average of Recall@10 and MRR on the development set from several runs as the result of each individual model. For model ensemble, we ensemble all the results from several runs from each model used for ensemble.

As shown in Table~\ref{tab:result:ubuntu}, for Ubuntu subtask 1, ESIM achieves 0.854 R@10 and 0.6401 MRR. If we remove the context composition (denoted by ``-CtxDec''), that is, removing context's local matching and matching composition, to accelerate the training process, R@10 and MRR drop to 0.845 and 0.6210, respectively. This result demonstrates the contribution of context composition to the ESIM model performance. Further discarding the last words instead of the preceding words for the context (denoted ``-CtxDec \& -Rev'') degrades R@10 and MRR to 0.840 and 0.6174. This comparison shows the efficacy of emphasizing the most recent context utterances in the ESIM model. Ensembling is performed by averaging output from models trained with different parameter initializations and different structures. Ensembling the above three models with several runs
(denoted ``Ensemble'') achieves significant improvement, reaching 0.887 R@10 and 0.6790 MRR. 

For Ubuntu subtask 2, the sentence-encoding based methods (denoted ``Sent-based'') achieves 0.082 R@10 and 0.0416 MRR. After ensembling several models with different parameter initializations (denoted ``Ensemble1''), R@10 and MRR are increased to 0.091 and 0.0475. Using ESIM to rerank the top 100 candidates predicted by ``Ensemble1'' improves R@10 and MRR significantly (0.125, 0.0713). Removing context's local matching and matching composition (``-CtxDec'') from the ESIM model degrades R@10 and MRR to 0.117 and 0.0620. Ensembling ESIM and -CtxDec models with several runs
(denoted ``Ensemble2'') again yields improvement (0.134 R@10,0.0770 MRR).

For Ubuntu subtask 4, we observe similar trend with subtask 1. ESIM achieves 0.887 R@10 and 0.6434 MRR, ``-CtxDec'' degrades performance to 0.877 R@10 and 0.6277 MRR, and ``-CtxDec \& -Rev'' further degrades performance to 0.875 R@10 and 0.6212 MRR. Ensembling the above three models with several runs 
(``Ensemble'') achieves 0.909 R@10 and 0.6771 MRR. 

For Ubuntu subtask 5, the dataset is the same as subtask 1 except for using the external knowledge of Linux manual pages. Concatenating pretrained word embeddings derived from Linux manual pages to the original five word embeddings followed by dimension reduction (``+W2V'') results in 0.858 R@10 and 0.6394 MRR, comparable with ESIM without exploring the external knowledge. However, ensembling the ensemble model for subtask 1 (0.887 R@10 and 0.6790 MRR, ``Ensemble1'') and the "+W2V" model with several runs 
brings further gain, reaching 0.890 R@10 and 0.6817 MRR (``Ensemble2'').

\begin{table}[ht]
\begin{center}
\scalebox{0.9}{
\begin{tabular}{l l r r r r}
\hline
\multicolumn{1}{l}{\textbf{Sub}} & \multicolumn{1}{l}{\textbf{Models}} & \multicolumn{1}{r}{\textbf{R@1}} & \multicolumn{1}{r}{\textbf{R@10}} & \multicolumn{1}{r}{\textbf{R@50}} & \multicolumn{1}{r}{\textbf{MRR}}  \\
\hline
\multirow{4}{*}{1} & ESIM & 0.534 & 0.854 & 0.985 & 0.6401  \\
& -CtxDec & 0.508 & 0.845 & 0.982 & 0.6210  \\
& -CtxDec \& -Rev  & 0.504 & 0.840 & 0.982 & 0.6174 \\
& Ensemble  & 0.573 & 0.887 & 0.989 & 0.6790  \\
\hline
\multirow{5}{*}{2} & Sent-based & 0.021  & 0.082 & 0.159 & 0.0416   \\
& Ensemble1 & 0.023 & 0.091 & 0.168 & 0.0475  \\
& ESIM & 0.043 & 0.125 & 0.191 & 0.0713   \\
& -CtxDec  & 0.034 & 0.117 & 0.191 & 0.0620 \\
& Ensemble2  & 0.048 & 0.134 & 0.194 & 0.0770  \\
\hline
\multirow{4}{*}{4} & ESIM & 0.515 & 0.887 & 0.988 & 0.6434  \\
& -CtxDec & 0.492 & 0.877 & 0.987 & 0.6277  \\
& -CtxDec \& -Rev  & 0.490 & 0.875 & 0.986 & 0.6212 \\
& Ensemble  & 0.551 & 0.909 & 0.992 & 0.6771  \\
\hline
\multirow{6}{*}{5} & -CtxDec \& -Rev  & 0.504 & 0.840 & 0.982 & 0.6174 \\
& -CtxDec & 0.508 & 0.845 & 0.982 & 0.6210  \\
& ESIM & 0.534 & 0.854 & 0.985 & 0.6401  \\
& Ensemble1  & 0.573 & 0.887 & 0.989 & 0.6790  \\
& +W2V & 0.530 & 0.858 & 0.986 & 0.6394  \\
& Ensemble2  & 0.575 & 0.890 & 0.989 & 0.6817  \\

\hline
\end{tabular}
}
\end{center}
\caption{Ablation analysis of removing context composition (-CtxDec), removing emphasizing most recent context utterances (-Rev), incorporating external domain knowledge (+W2V), and model ensemble (Ensemble) on the development set for the DSTC7 \textbf{Ubuntu} dataset. For subtask 5, ``+W2V" shows the results of concatenating the task specific word embeddings into the embedding combination. }
\label{tab:result:ubuntu}
\end{table}

Table~\ref{tab:result:advising} shows the same ablation analysis on the development set for the Advising dataset. We use ESIM without context's local matching and matching composition for computational efficiency. We observe similar trends on the Advising dataset as on the Ubuntu dataset. On subtask 1, subtask 3, and subtask 4, ``-CtxDec \& -Rev'' degrades R@10 and MRR over ``-CtxDec'', yet the ensemble of the two models with several runs produces significant gains over individual models, reaching 0.720 R@10 and 0.4010 MRR for subtask 1, 0.818 R@10 and 0.4848 MRR for subtask 2, and 0.760 R@10 and 0.4110 MRR for subtask 4. 
For Advising subtask 5, the dataset is the same as subtask 1 except for using the external knowledge of course information.
The ESIM model (i.e., with context's local matching and matching composition) achieves 0.662 R@10 and 0.3600 MRR, which has a small gain over  ``-CtxDec''. 
Ensembling the ensemble model for subtask 1 (0.720 R@10 and 0.4010 MRR,``Ensemble1'') and the ESIM model with several runs bring further gain, reaching 0.720 R@10 and 0.4115 MRR, denoted ``Ensemble2''.
To incorporate external knowledge from the course information, we extracted the Course, Course Title, and Description from the provided course knowledge file, trained word embeddings on these text, and concatenated the task specific embeddings with the five original word embeddings followed by dimension reduction for the ESIM model (denoted ``+W2V''). 
Different from the results from +W2V for Ubuntu subtask 5, for Advising subtask 5, +W2V brought a small gain on MRR, yet a comparable result on R@10. 
However, the ensemble of the ``Ensemble2'' model and the +W2V model with several runs, denoted ``Ensemble3'', still achieves a better performance, reaching 0.734 R@10 and 0.4199 MRR. 
In future work, we plan to continue investigating approaches for incorporating external domain knowledge in the ESIM model.

\begin{table}[ht]
\begin{center}
\scalebox{0.9}{
\begin{tabular}{l l r r r r}
\hline
\multicolumn{1}{l}{\textbf{Sub}} & \multicolumn{1}{l}{\textbf{Models}} & \multicolumn{1}{r}{\textbf{R@1}} & \multicolumn{1}{r}{\textbf{R@10}} & \multicolumn{1}{r}{\textbf{R@50}} & \multicolumn{1}{r}{\textbf{MRR}}  \\
\hline
\multirow{3}{*}{1} & -CtxDec & 0.222 & 0.656 & 0.954 & 0.3572  \\
& -CtxDec \& -Rev  & 0.214 & 0.658 & 0.942 & 0.3518 \\
& Ensemble  & 0.252 & 0.720 & 0.960 & 0.4010  \\
\hline
\multirow{3}{*}{3} & -CtxDec  & 0.320 & 0.792 & 0.978 & 0.4704 \\
& -CtxDec \& -Rev  & 0.310 & 0.788 & 0.978 & 0.4550 \\
& Ensemble  & 0.332 & 0.818 & 0.984 & 0.4848  \\
\hline
\multirow{3}{*}{4} & -CtxDec  & 0.248 & 0.706 & 0.970 & 0.3955  \\
& -CtxDec \& -Rev  & 0.226 & 0.714 & 0.946 & 0.3872  \\
& Ensemble  & 0.246 & 0.760 & 0.970 & 0.4110   \\
\hline
\multirow{7}{*}{5} & -CtxDec \& -Rev  & 0.214 & 0.658 & 0.942 & 0.3518 \\
& -CtxDec & 0.222 & 0.656 & 0.954 & 0.3572  \\
& Ensemble1  & 0.252 & 0.720 & 0.960 & 0.4010  \\
& ESIM & 0.224 & 0.662 & 0.946 & 0.3600 \\
& Ensemble2 & 0.262 & 0.720 & 0.960 & 0.4115 \\
& +W2V & 0.230 & 0.662 & 0.936 & 0.3662 \\
& Ensemble3 & 0.270 & 0.734 & 0.958 & 0.4199 \\
\hline
\end{tabular}
}
\end{center}
\caption{Same ablation analysis of removing context composition (-CtxDec), removing emphasizing most recent context utterances (-Rev), and model ensemble as in Table~\ref{tab:result:ubuntu}, but conducted on the development set for the DSTC7 \textbf{Advising} dataset. For subtask 5, ``+W2V" shows the results of adding the task specific word embeddings into the embedding combination. 
Note that, the official submission results in Table~\ref{tab:result} for subtask 5 of Advising is ``Ensemble1'' due to lack of enough time.}
\label{tab:result:advising}
\end{table}

\subsubsection{Effect of Combining Word Embeddings}

Table~\ref{tab:result:embedding:ubuntu} shows the ablation analysis of ESIM models using the different word embeddings listed in Table~\ref{tab:stat:word} and the ESIM using the concatenation of all of the word embeddings followed by dimension reduction for word representation. As can be seen from the table, the ESIM model using the glove.840B.300d embedding outperforms the ESIM models using the other embeddings. The ESIM model using the concatenation of all the embeddings followed by dimension reduction outperforms the ESIM models using each individual embedding, producing 1.0\% relative gain on R@10 and 1.3\% relative gain on MRR, compared to the ESIM model with the glove.840B.300d embedding.

\subsubsection{Effect of Data Augmentation and Sampling}

Table~\ref{tab:result:dataaugmentation-ratio} shows the ablation analysis of the heuristic data augmentation and tuning of the positive and negative sample ratio.  We find that for 1:1 as the positive and negative sample ratio, data augmentation produces 11.5\% relative gain on R@10 and 15.4\% relative gain on MRR; for 1:4, data augmentation produces 8.8\% relative gain on R@10 and 10.7\% relative gain on MRR. With data augmentation, using ratio 1:4 outperforms using ratio 1:1 by 2.3\% relative gain on R@10 and 3.8\% relative gain on MRR. 

Note that all results in Table~\ref{tab:result}, the results of our team (Team 3) in Table~\ref{tab:result:dstc7}, Table~\ref{tab:result:ubuntu} and Table~\ref{tab:result:advising} (with the exception of
``ESIM'' and ``+W2V'' and their ensembles for Advising Subtask5 in Table~\ref{tab:result:advising}) are from our implementation based on Theano. Due to the change in the Theano environment (probably due to some updates of GPU drivers) after the challenge, and due to the ending of Theano's active development and support, we made another implementation based on TensorFlow to benefit from TensorFlow's active development and support. We conducted comparison with previous work (Table~\ref{tab:result:public}) and the ablation analysis for the ESIM models, including combining word embeddings (Table~\ref{tab:result:embedding:ubuntu}), data augmentation and sampling (Table~\ref{tab:result:dataaugmentation-ratio}), and ``ESIM'' and ``+W2V'' and their ensembles for Advising Subtask5 in Table~\ref{tab:result:advising}, using the TensorFlow based implementation. Note that there are some differences on the ESIM model results for Ubuntu subtask 1 development set between the Theano implementation and the TensorFlow implementation. R@1, R@10, R@50, and MRR from the TensorFlow based implementation are 0.518, 0.850, 0.985, and 0.6287; these results from the Theano version are 0.534, 0.854, 0.985, and 0.6401. Note that every ablation analysis in this paper is conducted on one consistent codebase to facilitate fair comparisons within each table (with the only exception of ``ESIM'' and ``+W2V'' and their ensembles for Advising Subtask5 in Table~\ref{tab:result:advising} versus the rest of Table~\ref{tab:result:advising}), hence our observations are valid.

\begin{table}[ht]
\begin{center}
\scalebox{0.9}{
\begin{tabular}{l r r r r}
\hline
\multicolumn{1}{l}{\textbf{Embedding}} & \multicolumn{1}{r}{\textbf{R@1}} & \multicolumn{1}{r}{\textbf{R@10}} & \multicolumn{1}{r}{\textbf{R@50}} & \multicolumn{1}{r}{\textbf{MRR}}  \\
\hline
glove.6B.300d	& 0.432	& 0.773	& 0.974	& 0.5443 \\
glove.840B.300d	& 0.514	& 0.842	& 0.985	& 0.6215 \\
glove.twitter.27B.200d	& 0.415	& 0.754	& 0.968	& 0.5252 \\
wiki-news-300d-1M.vec	& 0.474	& 0.820	& 0.979	& 0.5886 \\
crawl-300d-2M.vec	& 0.503	& 0.827	& 0.981	& 0.6114 \\
\hline
Combination & \textbf{0.518}	& \textbf{0.850}	& \textbf{0.985}	& \textbf{0.6287} \\
\hline
\end{tabular}
}
\end{center}
\caption{Ablation analysis of using different word embeddings, compared with combining all five word embeddings followed by dimension reduction as in the submitted system, on the subtask 1 development set for the DSTC7 \textbf{Ubuntu} dataset. }
\label{tab:result:embedding:ubuntu}
\end{table}

\begin{table}[ht]
\begin{center}
\scalebox{0.9}{
\begin{tabular}{l r r r r}
\hline
\multicolumn{1}{l}{\textbf{Models}} & \multicolumn{1}{l}{\textbf{R@1}} & \multicolumn{1}{r}{\textbf{R@10}} & \multicolumn{1}{r}{\textbf{R@50}} & \multicolumn{1}{r}{\textbf{MRR}}  \\
\hline
Data Augmentation, 1:4	& \textbf{0.518}	& \textbf{0.850}	& \textbf{0.985}	& \textbf{0.6287} \\
Data Augmentation, 1:1	& 0.495	& 0.831	& 0.982	& 0.6063 \\
No Data Augmentation, 1:4	& 0.463	& 0.781	& 0.973	& 0.5679 \\
No Data Augmentation, 1:1	& 0.415	& 0.745	& 0.972	& 0.5247 \\
\hline
\end{tabular}
}
\end{center}
\caption{Ablation analysis of using the heuristic data augmentation or not, and comparing the positive:negative sample ratio 1:4 or 1:1, on the subtask 1 development set for the DSTC7 \textbf{Ubuntu} dataset. }
\label{tab:result:dataaugmentation-ratio}
\end{table}

\subsection{BERT Model Results and Ablation Analysis}
\label{subsect:bert}
After the DSTC7 challenge, we also investigate the efficacy of applying the Bidirectional Encoder Representations from Transformers (BERT) model~\citep{DBLP:journals/corr/abs-1810-04805}, which has created state-of-the-art models for a wide variety of NLP tasks, for multi-turn response selection. 

The BERT model is a multi-layer bidirectional Transformer encoder based on the original Transformer model~\citep{DBLP:conf/nips/VaswaniSPUJGKP17}. 
The input representation is created by summing the corresponding 
WordPiece embeddings~\citep{DBLP:journals/corr/WuSCLNMKCGMKSJL16}, position embeddings, and segment embeddings. A special classification embedding ([CLS]) is inserted as the first token and a special token ([SEP]) is added as the final token. Given an input token sequence $[x]_1^T$, the output of BERT is $[h]_1^T$.

The BERT model is pretrained with two training tasks on large-scale unlabeled text, i.e., masked language model and next sentence prediction. The pretrained BERT model provides a powerful context-dependent sentence representation and can be used for various target tasks, e.g., multi-turn response selection, through the fine-tuning procedure, similar to how it is used for other NLP tasks. For fine-tuning for multi-turn response selection, we compose the input to the BERT model by concatenating the context and the response, using the same two special tokens, \_\_eou\_\_ and \_\_eot\_\_, where \_\_eou\_\_ denotes end-of-utterance and \_\_eot\_\_ denotes end-of-turn.
The BERT fine-tuning data is the same as the data used for our ESIM model training, that is, the data has been augmented using a similar data augmentation strategy as in~\citep{DBLP:conf/sigdial/LowePSP15} and tuned for the ratio between positive and negative responses.

We use the pretrained English uncased BERT-base model\footnote{https://github.com/google-research/bert}, which has 12 layers, 768 hidden states, and 12 heads. The BERT-base model is pretrained on BooksCorpus (800M words)~\citep{DBLP:conf/iccv/ZhuKZSUTF15} and English Wikipedia (2,500M words). For fine-tuning, all hyperparameters are tuned on the development set. The maximum sequence length is 128. The batch size is 32. Adam~\citep{DBLP:journals/corr/KingmaB14} is used for optimization. We compared the initial learning rate among [1e-5, 2e-5, 3e-5, 5e-5]. The dropout probability is 0.1. The number of training epochs is 2 for the Ubuntu dataset and 1 for the Advising1 and Advising2 datasets.

\begin{table}[ht!]
\begin{center}
\scalebox{0.9}{
\begin{tabular}{l r r r r }
\hline
\multicolumn{1}{l}{\textbf{Models}} & \multicolumn{1}{r}{\textbf{R@1}} & \multicolumn{1}{r}{\textbf{R@10}} & \multicolumn{1}{r}{\textbf{R@50}} & \multicolumn{1}{r}{\textbf{MRR}}  \\
\hline
Our submitted ESIM & 0.645 &	0.902 &	\textbf{0.994} &	0.7350 \\
\hline
Vig and Ramea's MT-EE &	0.478 &	0.765 &	0.952 &	0.578 \\
Vig and Ramea's GPT	& 0.489	& 0.799	& 0.972	& 0.595 \\
Vig and Ramea's BERT &	0.530 &	0.817 &	0.978 &	0.632 \\ 
\hline 
BERT lr1e-5	& 0.661	& 0.892	& 0.981	& 0.7410 \\
BERT lr2e-5	& 0.663	& 0.905	& 0.987	& 0.7522 \\
BERT lr3e-5	& \textbf{0.692}	& \textbf{0.913}	& 0.986	& \textbf{0.7680} \\
BERT lr5e-5	& 0.652	& 0.900	& 0.991	& 0.7402 \\
\hline
BERT lr1e-5, No Data Augmentation, 1:1 & 0.553 & 0.812 & 0.966 & 0.6427 \\
BERT lr2e-5, No Data Augmentation, 1:1 & 0.562 & 0.820 & 0.973 & 0.6528 \\
BERT lr3e-5, No Data Augmentation, 1:1 & 0.561 & 0.839 & 0.972 & 0.6578 \\
BERT lr5e-5, No Data Augmentation, 1:1 & 0.544 & 0.827 & 0.984 & 0.6467 \\
\hline
BERT no-pre-train lr2e-5 & 0.301 & 0.628 & 0.918 & 0.4138 \\
\textbf{Mean} & 0.297 & 0.608 & 0.909 & 0.4054 \\
\textbf{Standard Deviation} & 0.100 & 0.077 & 0.019 & 0.0925 \\
\hline
\end{tabular}
}
\end{center}
\caption{Our post-DSTC7 BERT results (the third group) on the hidden test sets for the DSTC7 response selection challenge \textbf{Ubuntu} data, compared to our submitted ESIM ensemble results (as in Table~\ref{tab:result}). Unless noted otherwise, the pretrained BERT model was fine-tuned with 2 epochs on the same training data as the ESIM model for the Ubuntu data set, i.e., with the heuristic data augmentation and 1:4 for positive:negative samples. Results are shown for using the initial learning rates (lr) from [1e-5, 2e-5, 3e-5, 5e-5]. The second group results, including Multi-turn ESIM + ELMo (denoted MT-EE), OpenAI GPT, and BERT model results, are all cited from~\citep{Vig19}. The fourth group of results are from the pretrained BERT model fine-tuned with 2 epochs, but on data without the heuristic data augmentation and with 1:1 for positive:negative samples. The fifth group of results is from the BERT model without pre-training, trained on the same training data as the third group. Note that we only present the best results here for BERT no-pre-train, which is from using lr2e-5, by comparing using lr from [1e-5, 2e-5, 3e-5, 5e-5]. For lr2e-5, we ran the experiment five times and present the best results together with the mean and standard deviation of results from these five runs.}
\label{tab:bert-ubuntu-result}
\end{table}

\begin{table}[ht]
\begin{center}
\scalebox{0.79}{
\begin{tabular}{l c c c c c c c c}
\hline
\multicolumn{1}{l}{\textbf{Models}} & \multicolumn{4}{c}{\textbf{Advising1}} & \multicolumn{4}{c}{\textbf{Advising2}} \\
 & \multicolumn{1}{c}{\textbf{R@1}} & \multicolumn{1}{c}{\textbf{R@10}} & \multicolumn{1}{c}{\textbf{R@50}} & \multicolumn{1}{c}{\textbf{MRR}}
& \multicolumn{1}{c}{\textbf{R@1}} & \multicolumn{1}{c}{\textbf{R@10}} & \multicolumn{1}{c}{\textbf{R@50}} & \multicolumn{1}{c}{\textbf{MRR}}
\\
\hline
Our submitted ESIM	& 0.398	& 0.844	& \textbf{0.986}	& 0.5408	& 0.214	& 0.630	& \textbf{0.948}	& 0.3390 \\
\hline
Vig and Ramea's MT-EE & - & - & - & - & 0.132	& 0.512	& 0.890	& 0.252 \\
Vig and Ramea's GPT	& - & - & - &- & 0.172	& 0.568	& 0.932	& 0.293 \\
Vig and Ramea's BERT & - & - & - & - & 0.186	& 0.580	& 0.942	& 0.312 \\
\hline
BERT lr1e-5 & 0.671	& 0.874	& 0.984	& 0.7431	& 0.250	& 0.630	& 0.934	& 0.3732 \\
BERT lr2e-5 & 0.681	& 0.880	& 0.972	& 0.7486	& 0.232	& 0.644	& 0.916	& 0.3620 \\
BERT lr3e-5 & 0.689	& \textbf{0.890}	& 0.978	& 0.7530	& 0.252	& \textbf{0.682}	& 0.932	& 0.3831 \\
BERT lr5e-5 & 0.659	& 0.884	& 0.982	& 0.7320	& 0.228	& 0.614	& 0.938	& 0.3507 \\
\hline
BERT lr2e-5 + course info & \textbf{0.695} & 0.876 & 0.980 & \textbf{0.7564} & \textbf{0.262} & 0.668 & 0.942 & \textbf{0.3966} \\
BERT lr3e-5 + course info & 0.683 & 0.874 & 0.982 & 0.7501 & 0.246 & \textbf{0.682} & 0.942 & 0.3876 \\
\hline
\end{tabular}
}
\end{center}
\caption{Our post-DSTC7 BERT results (the third group) on the hidden test sets for the DSTC7 response selection challenge \textbf{Advising} data, compared to our submitted ESIM ensemble results (as in Table~\ref{tab:result}). The BERT model was trained with 1 epoch. Results are shown for using the initial learning rates from [1e-5, 2e-5, 3e-5, 5e-5]. The second group of results, including Multi-turn ESIM + ELMo (MT-EE), OpenAI GPT, and BERT model results, are cited from~\citep{Vig19}. Note that~\citep{Vig19} only reported results on the Advising2 test set. The fourth group of results is from incorporating suggested course information into the BERT model.}
\label{tab:bert-advising-result}
\end{table}

The results on the Ubuntu and Advising1, Advising2 test sets are shown in Table~\ref{tab:bert-ubuntu-result} and Table~\ref{tab:bert-advising-result}, respectively.  In both tables, the ESIM\_submitted model is our submitted ESIM ensemble model, the results of which rank the best on 7 subtasks, rank the second best on subtask 2 of Ubuntu, and overall rank top 1 on both datasets of the DSTC7 response selection challenge. The following observations are made from Table~\ref{tab:bert-ubuntu-result} and Table~\ref{tab:bert-advising-result}.

\paragraph{\textbf{Overall BERT performance}} On both Ubuntu and Advising1 and Advising2 test sets, the BERT model with initial learning rate 3e-5 achieves the best results, yielding an absolute gain of 3.3\% on MRR on the Ubuntu test set, and an absolute gain of 4.4\% on MRR on the Advising2 data set, compared to our ESIM\_submitted model. Similar to the ESIM results in Table~\ref{tab:result}, the BERT model results on the Advising1 test set are better than those on the Advising2 test set, because test case 1 (Advising1) has some dependency on the training dataset.

\paragraph{\textbf{Effect of heuristic data augmentation and sampling}} Our BERT model results are much better than those reported in~\citep{Vig19}. \cite{Vig19} used the BERT-base, cased model and the overall sequence length is set to 512 tokens including both context and response. In comparison, we used the BERT-base, uncased model and the overall sequence length is set to 128. We hypothesize that our much better BERT model results are probably due to our data augmentation strategy and tuning of the ratio between positive and negative responses. To verify this hypothesis, we ran another set of experiments for BERT models with the same initial learning rates, without the heuristic data augmentation and using 1:1 as the ratio between positive and negative samples (the fourth group of results in Table~\ref{tab:bert-ubuntu-result}). The best results are from setting the initial learning rate as 3e-5. However, R@1, R@10, R@50, and MRR now are only slightly better than those BERT results reported by~\citep{Vig19}. These results verified our hypothesis that the heuristic data augmentation and tuning of the positive/negative sample ratio that we employed have a significant efficacy on the BERT model performance for response selection.

\paragraph{\textbf{Effect of pre-training}} We also investigate the efficacy of BERT pre-training for the response selection task, by randomly initializing the BERT model (i.e., without pre-training) and then training 2 epochs for response selection. The fifth group of results in Table~\ref{tab:bert-ubuntu-result} shows the results from this setting on the Ubuntu test set. The best results in this setting are from using the initial learning rate 2e-5, which are significantly worse than those results in the third group (note that the training data for the BERT models without pre-training are the same as used by the third group of BERT models with pre-training). To further verify whether the BERT model results without pre-training are consistently low, we conduct 5 runs of randomly initializing the BERT model with the initial learning rate 2e-5. The mean and standard deviation of the results of the 5 runs are shown in Table~\ref{tab:bert-ubuntu-result}.
The mean shows a much lower performance compared to the BERT model results with pre-training (i.e., the third group of results in Table~\ref{tab:bert-ubuntu-result}); and the high standard deviation shows the trained models with random initialization are quite unstable. These results demonstrate that removing the BERT pre-training is detrimental to the BERT model training stability and performance on the response selection task. We also observe significant performance degradation from randomly initialized BERT models on both Advising1 and Advising2 test sets, compared to the BERT models with pre-training.

\paragraph{\textbf{Effect of incorporating external domain knowledge}} We investigate the effectiveness of incorporating external domain knowledge in the BERT model on the Advising1 and Advising2 datasets.
The fourth group of results in Table~\ref{tab:bert-advising-result} shows the BERT model results after incorporating the external domain knowledge by concatenating a natural language sentence representation for the course information provided in the knowledge base using a pre-defined template as in~\citep{ganhotra2019knowledge}.
An example natural language sentence representation for a suggested course is: ``\textit{EECS376} is \textit{Foundations of Computer Science} , has \textit{easy} workload , \textit{medium} class size , \textit{4} credits , has a \textit{discussion}''.
We observe that incorporating external domain knowledge with this approach brings gain for the learning rate 2e-5 and 3e-5 (lr2e-5 and lr3e-5) over the BERT model results in the third group for Advising2. In future work, we plan to investigate other approaches for incorporating external domain knowledge and knowledge representations for the BERT model.

\paragraph{\textbf{Comparing ESIM and BERT models}} The significant gain from the BERT models over our submitted ESIM models is probably due to the following reasons. The BERT model is pretrained on a large amount of unlabeled text in an end-to-end manner, uses self-attention mechanism that attends the entire context, and the BERT model is jointly trained on two tasks, masked language model (MLM) and next sentence prediction (NSP). The NSP task predicts whether the second sentence in an input sentence pair follows the first, which is highly related to the response selection task. It is verified in earlier research that the NSP pre-training task is important for a BERT model on the QNLI answer selection task, a task highly related to the response selection task.

On the other hand, the ESIM models achieve an outstanding performance on the noetic end-to-end response selection task with a significant advantage on computational complexity compared to the BERT models. For training, on the DSTC7 Advising subtask 1, using a single NVIDIA Tesla V100 GPU card, training the ESIM model with 1 epoch, with the maximum sequence length being 330 (300 for the context + 30 for the response), takes 2.5 hours; in contrast, fine-tuning the BERT model with 1 epoch, with the maximum sequence length being 128, takes 23 hours. For inference, the computational complexity of the ESIM model is linear w.r.t. the number of tokens; in comparison, the BERT model has self-attention layers that have a computational complexity being quadratic to the sequence length, i.e., the number of tokens. On the DSTC7 Advising subtask1 development set, using a single NVIDIA Tesla V100 GPU card, the ESIM model inference takes 1 minute, whereas the BERT model inference takes 11 minutes.

\section{Conclusion}
Previous state-of-the-art multi-turn response selection models used hierarchy-based (utterance-level and token-level) neural networks  to explicitly model the interactions among the different turns' utterances for context modeling. In this paper, we demonstrated that a sequential matching model based only on chain sequence can outperform all previous models, including the previous sentence-encoding based models, the previous sequence-based matching models, and the previous hierarchy-based methods, suggesting that the potentials of such sequential matching approaches have not been fully exploited in the past. Specially, the proposed model achieved top 1 results on both datasets under the noetic end-to-end response selection challenge in DSTC7, and yielded new state-of-the-art performances on two large-scale public multi-turn response selection benchmarks over all previous models, including the state-of-the-art hierarchy-based models.

Future work includes the following directions: (1) We plan to explore approaches for effectively representing and incorporating external knowledge~\citep{DBLP:conf/acl/InkpenZLCW18} in the ESIM model and the BERT model, such as knowledge graph and user profile. It is important to advance the understanding of how to effectively represent the interactions between the context and the external knowledge for the response selection task. (2) We plan to investigate the efficacy of the ESIM model on multi-turn response selection for multi-domain multi-turn dialogues or even multi-modal dialogues. (3) We plan to investigate the effectiveness of applying the ESIM model for quality assessment for dialogue data collection.





\bibliographystyle{elsarticle-harv}
\bibliography{ref}




\end{document}